\documentclass[research]{fcs}
\newcommand{\shortauthor}[1]{}

\usepackage[utf8]{inputenc} 
\usepackage[T1]{fontenc}    

\usepackage{hyperref}       
\usepackage{url}            
\usepackage{booktabs}       
\usepackage{amsfonts}       
\usepackage{nicefrac}       
\usepackage{microtype}      
\usepackage{caption}        
\usepackage{listings}       
\usepackage{xcolor}         

\usepackage{amsmath}

\usepackage{amscd,amsfonts,amsbsy}
\usepackage{balance}
\usepackage{graphicx}
\usepackage{subcaption}
\usepackage{multirow}
\usepackage{latexsym,bm}
\usepackage{mathtools}
\usepackage{euscript}

\usepackage{makecell}
\usepackage{textcomp}

\usepackage{enumitem}
\setenumerate[1]{itemsep=0pt,partopsep=0pt,parsep=\parskip,topsep=5pt}
\setitemize[1]{itemsep=0pt,partopsep=0pt,parsep=\parskip,topsep=5pt}
\setdescription{itemsep=0pt,partopsep=0pt,parsep=\parskip,topsep=5pt}

\usepackage{inconsolata}
\usepackage{mdframed}
\usepackage{xspace}

\usepackage{natbib}

\usepackage{svg}

\newcommand{\minisection}[1]{\vspace{0pt}\noindent\textbf{#1.}}

\newcommand{\modelname}{FleetPilot\xspace}

\title{Fleet Pilot: Fine-Grained Steering Empowers \\ Real-World LLM Personalization}

\shorttitle{\modelname}
\shortauthor{Kounianhua et al.}

\author{Kounianhua Du$^1$, Jianxing Liu$^1$, Kangning Zhang$^1$, 
 Wenxiang Jiao$^2$, Yuan Lu$^2$, \\
 Jiarui Jin$^2$, Jianghao Lin$^1$, Weiwen Liu$^1$, Yong Yu$^1$, Weinan Zhang$^1$}
\address{$^1$Shanghai Jiao Tong University, $^2$ Xiaohongshu Inc., Shanghai, China}
\fcssetup{
  received       = {},
  accepted       = {},
  corr-email     = {kounianhuadu@sjtu.edu.cn},
}

\begin{abstract}
With the rapid evolution of large language models (LLMs), there is an increasing demand for effective personalization techniques that can adapt model behavior to individual user preferences. Personalized large language models (LLMs) must infer a user's query-dependent preferences from limited, heterogeneous, and evolving interaction histories. Existing approaches rely primarily on in-context personalization, parametric adaptation, or personalized reward modeling, but respectively incur recurring costs in context management, model optimization and deployment, or multi-candidate generation and scoring. These limitations become particularly restrictive when user observations are sparse and the preference evidence relevant to each query varies across contexts and over time.

To address these challenges, we propose \modelname, a fine-grained and instance-tailored steering framework that dynamically generates sample-level interference vectors from user data and injects them into the model's forward pass for personalized adaptation. 
 Our approach introduces: match contrast that produces user-relevant signals, a fine-grained steering component that captures nuanced signals by hooking activations from attention and MLP layers, and a query-conditioned aggregation module that synthesizes these signals into contextually relevant enhancements. The method demonstrates high flexibility and data efficiency, excelling in fast-changing distribution and high data sparsity scenarios. 
 
 Extensive experiments across diverse scenarios, including short-to-long text generation, and web function calling, validate the effectiveness and compatibility of \modelname. Results show that our method  consistently brings significant performance gains, and show extreme superiority in fast-shifting  and data-sparse environments. Additionally, \modelname rids us of training cycles and incurs light overhead of storage or inference latency. These make \modelname a strong and real-world deployable method.
 Implementation is available \footnote{https://anonymous.4open.science/r/FleetPilot}. 
 
\end{abstract}
\keywords{LLM, Personalization, Steering}

\begin{document}
\section{Introduction}
\label{sec:intro}
The advent of large language models (LLMs)~\cite{dubey2024llama,yang2025qwen3} has catalyzed transformative advances across a multitude of domains. A particularly compelling frontier within this landscape is the pursuit of personalization—the ability to tailor model behavior and outputs to individual user preferences~\cite{liu2025survey,cai2025large,wang2025never,salemi2023lamp,kumar2024longlamp}. Achieving effective personalization is paramount for deploying LLMs in real-world applications such as personal assistant~\cite{wang2025roles,guha2015user}, conversational agents~\cite{yusuf2025pedagogical,liu2025proactive}, and adaptive learning systems~\cite{iqbal2025personalized,saleem2025ai}. 

\begin{figure*}
    \centering
    \includegraphics[width=1.0\linewidth]{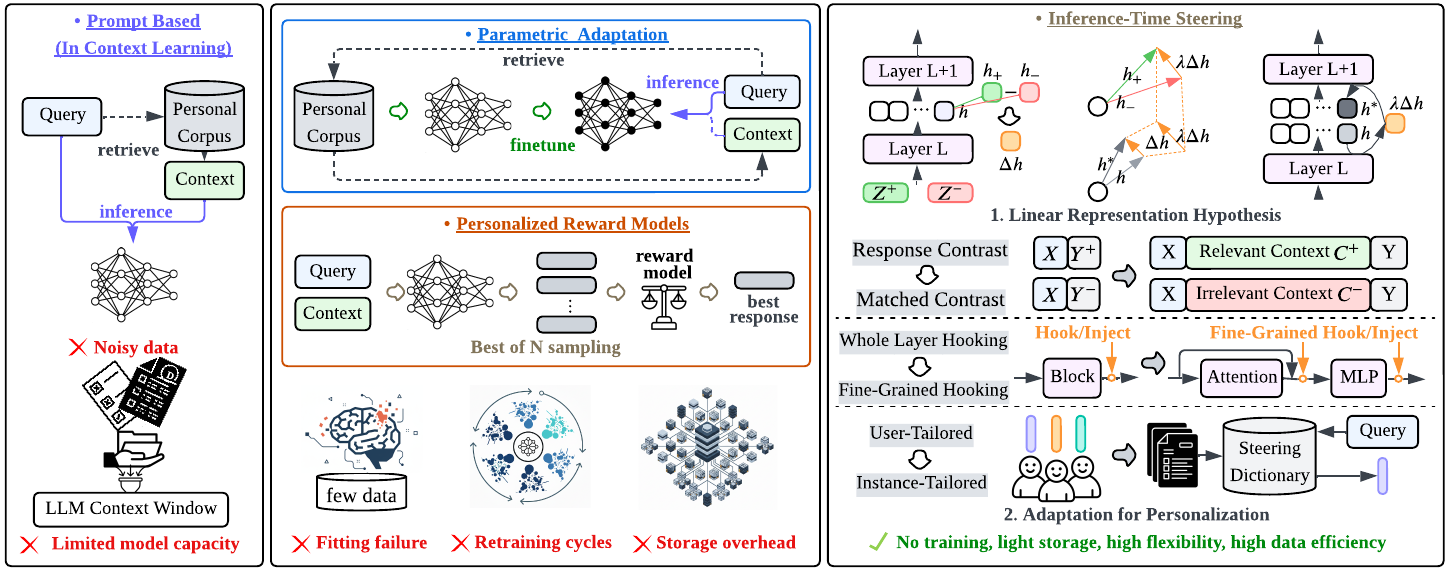}
    \caption{Personalization paradigms and the design of \modelname. Left: prompt-based personalization supplies retrieved history through a context window, where noisy evidence and limited model capacity constrain its use. Middle: parametric adaptation and personalized reward modeling introduce optimization or candidate-scoring stages, with challenges in sparse-data fitting, model updates, storage, and best-of-$N$ inference. Right: inference-time steering modifies intermediate activations in a frozen LLM, motivated by the linear representation hypothesis. \modelname adapts this principle to personalization through three design changes: response contrast to matched contrast, whole-layer hooking to fine-grained hooking, and user-tailored to instance-tailored steering.
    }
    \vspace{-10pt}
    \label{fig:intro}
\end{figure*}

A practical personalization system must infer a user's current, query-dependent preference from a limited and continuously evolving interaction history. Formally, for user $u$ at time $t$, let $\mathcal{H}_u^{t}=\{(q_i,y_i)\}_{i<t}$ denote the observations available before a new query $q_t$. The system must generate an output aligned with the latent preference $p_u^t(y\mid q_t)$, although this distribution is only partially observed, may vary with the query, and may change over time. 
Production environments expose challenging regimes:
\begin{itemize}[leftmargin=*]
\item \textbf{Limited Preference Observations (Data Scarcity).} For new or infrequent users, $|\mathcal{H}_u^{t}|$ can be extremely small, often fewer than ten interactions. The resulting cold-start regime~\cite{zhang2025cold} provides weak and noisy evidence about individual preferences. An effective system must therefore achieve high statistical efficiency rather than rely on abundant user-specific supervision.

\item \textbf{Non-Stationary Preferences (Temporal Adaptability).} The target distribution is generally time-dependent, i.e., $p_u^{t+1}(y\mid q)\neq p_u^{t}(y\mid q)$ after a change in interests, goals, or context~\cite{liu2024retrieval}. Since such changes may occur within days or even hours, personalization must incorporate recent evidence at inference time; adaptation tied to periodic optimization and redeployment can leave the system aligned with stale preferences.


\item \textbf{Context-Dependent Preference Heterogeneity (Instance Specificity).} A single user can exhibit different preferences across queries, domains, and output forms, from short responses and long-form generation to structured tool calls~\cite{jiang2024importance}. Consequently, the relevant signal is not fully captured by one static, user-level representation. The system must identify which subset of $\mathcal{H}_u^{t}$ is pertinent to $q_t$ and apply the corresponding preference without allowing unrelated or conflicting history to dominate.

\end{itemize}

These challenges are coupled: scarce observations make drift difficult to estimate, while temporal and contextual heterogeneity further reduce the amount of history relevant to any individual query. We therefore require a practical method to be \emph{data-efficient}, \emph{rapidly updateable}, and \emph{instance-specific}, while avoiding repeated training or multi-candidate inference.

Existing methodologies for LLM personalization can be broadly categorized into three paradigms. 
\textbf{Prompt-based} methods leverage  in-context-learning (ICL)~\cite{dong2022survey}  ability of llms,  strategically inserting user profiles or historical interactions into the LLM context window for personalized output~\cite{madaan2022memory,CFRAG,qian2024memorag}. 
\textbf{Parametric adaptation} methods include One-Peft-All-User~\cite{lin2024rella} that learns collaborative signals within the whole dataset by tuning LLMs on all the users' data and One-Peft-Per-User~\cite{tan2024democratizing,tan2024personalized,PLoRA} that performs parameter-efficient fine-tuning on user-specific data, tailoring user-specific weights to offer personalization.
\textbf{Personalized reward modeling} methods learn preference-aware scoring functions that evaluate multiple candidate generations and select the output most aligned with the target user~\cite{zhuang2024hydra,chen2025pal,ryan2025synthesizeme}. 


Despite their effectiveness, the three paradigms expose different bottlenecks under these real-world requirements.

\textbf{Prompt-based personalization} is training-free and immediately updateable, but its performance is governed by the quality of context content used and the llm capacity~\cite{amazon2025context}.  Long, noisy records can introduce stale, contradictory, or irrelevant evidence that distracts the generator; truncation or profile summarization can instead discard minority and recently emerging preferences. Longer prompts also increase prefill latency and repeatedly expose raw personal records to the inference service~\cite{liu2025survey,vuong2021does}. \citep{lin2024rella} shows that these methods often reach their peak performances at limited context tokens, even far less than the llm capability upper bound.


\textbf{Parametric adaptation} encodes preferences more compactly, but converts each preference update into an optimization and deployment problem. A shared adapter must reconcile heterogeneous users and can suffer negative transfer, whereas one-adapter-per-user systems require user-specific supervision and incur training, storage, versioning, and adapter-switching costs that grow with the user population~\cite{liu2025survey,lin2024rella,tan2024democratizing,tan2024personalized}. With only a few interactions, parameter estimates are unstable and prone to fitting incidental features; once preferences drift, stale adapters must be retrained or replaced rather than conditioned on the latest interaction at inference time.

\textbf{Personalized reward modeling} moves personalization into an explicit scoring function, yet sparse pairwise feedback underdetermines a user's context-dependent utility and may encourage reliance on spurious correlates. Models learned from pooled feedback can blur conflicting tastes, while per-user models face a severe bias--variance trade-off when feedback is scarce~\cite{zhuang2024hydra,chen2025pal,ryan2025synthesizeme}. Moreover, reranking and rejection-sampling pipelines can only choose among generated candidates: their quality is bounded by candidate coverage, and increasing that coverage through best-of-$N$ or rejection sampling multiplies generation and scoring cost~\cite{gui2024bonbon,liu2023statistical}. A reward function estimated offline can also become miscalibrated as preferences or task distributions change, requiring fresh feedback and model updates.

\smallskip



\smallskip

Activation steering~\cite{rimsky2024steering,turner2023steering,zou2023representation} has been used in toxicity reduction and deception control, where high-level behavioral attributes are hypothesized to be encoded as approximately linear directions in the model's residual-stream activation space (the linear representation hypothesis), so that a single contrastive difference vector suffices to identify each concept. Under this hypothesis, adding or subtracting a scaled version of that direction at an appropriate mid-to-late layer produces a monotonic, causally faithful shift of the corresponding behavior at inference time.
It offers a complementary interface for personalization, which is attractive because it allows behavioral adaptation without gradient-based optimization or per-user model checkpoints. However, applying it to realistic personalization requires more than a fixed behavioral direction. Sparse and heterogeneous histories provide only indirect evidence of user preferences, and the evidence relevant to one query may be inappropriate for another. The central challenge is therefore to extract useful personalization signals, preserve their effects at appropriate computational interfaces, and select the signals relevant to the current query.

To address this challenge, we propose \modelname, a fine-grained, instance-tailored steering framework built around three coordinated designs.

\textbf{Matched contrast} varies the conditioning context while holding a historical query and its recorded response fixed. By contrasting activations under user-relevant and user-irrelevant contexts, it captures changes associated with personalized conditioning without introducing differences in response content.

\textbf{Fine-grained hooking} constructs a pair of steering directions for each historical interaction by contrasting the model's activations under user-relevant and user-irrelevant contexts. Rather than collapsing the contrastive signal into a single post-block direction, it separately captures the post-attention residual and the MLP output. These two hook points expose complementary computational pathways and preserve distinct preference-induced effects that whole-layer steering may conflate.

\textbf{Query-conditioned aggregation} then adapts these historical directions to the current query. At inference time, \modelname retrieves the most semantically relevant interactions from the user's steering-vector dictionary and aggregates their attention- and MLP-side directions separately. The resulting pair of query-specific directions is applied through position-matched sequential injection, enabling the model to condition on the most pertinent preference evidence in a single auto-regressive generation. This design supports immediate adaptation to sparse observations and evolving interests without gradient updates or per-user model deployment.

Overall, \modelname constructs a pair of query-specific interference vectors per user query by aggregating contextually relevant steering vectors, which are prebuilt from fine-grained contrasting model's internal activations under the positive (personal) prompts and the negative (impersonal) prompts.
Because the base model remains frozen and the query-specific steering directions are injected only transiently into intermediate activations, \modelname requires neither gradient-state storage nor per-user model checkpoints, while introducing only modest inference-time latency. Empirical studies show that \modelname outperforms all the competing baselines across different validation scenarios and show extreme superiority in dynamic and data-sparse environments.

\smallskip
\noindent\textbf{Contributions} are summarized below:
\begin{itemize}[leftmargin=*]
  \item \textbf{Comprehensive summarization of existing personalization paradigms and their fragility towards realistic challenges. Point out the correctness of inference-time steering as a training-free, flexible, robust, and low-cost paradigm.} We analyze existing personalization paradigms, including prompt-based conditioning, parametric adaptation, and personalized reward modeling, through the lens of data scarcity, temporal preference drift, and query-dependent preference heterogeneity. This analysis identifies their respective limitations and motivates inference-time activation steering as a training-free, flexible, and resource-efficient interface for personalization, establishing the design requirements for robust adaptation under realistic conditions.

  \item \textbf{Methodology novelty:} 
  We propose \modelname, which adapts activation steering to personalization through three coordinated designs. First, matched contrast holds historical queries and responses fixed while varying user-relevant and user-irrelevant contexts, extracting steering signals associated with personalized conditioning. Second, fine-grained hooking preserves separate directions from post-attention residuals and MLP outputs, while sequential dual-injection applies them at their corresponding computational interfaces for nuanced control. Third, query-conditioned retrieval and aggregation move beyond fixed concept-, behavior-, or user-level directions to instance-tailored interventions, flexibly selecting the historical signals relevant to each query without updating model parameters.

  \item \textbf{Empirical validation and analysis}: Experiments on various datasets including short content generation, long content generation, and personal web function calling demonstrate \modelname’s consistent superiority over all competing baselines. Particularly, \modelname excels in data-scarce scenario and fast-shifting distribution, with light overhead and no training cost. These justify \modelname as a robust and real-world deployable method.
  


\end{itemize}

\section{Related Work}
\label{sec:related_work}

\subsection{Prompt-Based Personalization}
Prompt-based methods leverage the in-context learning capabilities of LLMs to personalize generation without changing model parameters. They retrieve relevant records from a user's personal corpus or construct profiles that summarize historical preferences, and supply this information alongside the current query. LaMP~\cite{salemi2023lamp} studies retrieval-augmented personalization across language tasks, while profile-based approaches compress interaction histories into more concise preference descriptions~\cite{cai2025large}. Memory and retrieval frameworks further investigate how to organize and retrieve contextual evidence~\cite{madaan2022memory,CFRAG,qian2024memorag}.

These methods are flexible and can incorporate new records without retraining. Their effectiveness nevertheless depends on retrieval quality, context composition, and the model's ability to use the supplied evidence. Finite context windows constrain how much history can be included, and adding noisy or weakly relevant records can reduce performance even within that limit~\cite{vuong2021does,lin2024rella,amazon2025context}. \modelname is complementary to this paradigm: retrieved text supplies explicit evidence, while query-conditioned steering supplies an additional activation-space intervention.

\subsection{Parametric Adaptation}
Parametric personalization incorporates user information through model optimization, commonly using parameter-efficient fine-tuning such as Low-Rank Adaptation (LoRA)~\cite{hu2022lora}. One line of work trains a shared adapter using multiple users' histories, allowing the model to learn collaborative signals and task-specific response patterns. ReLLa~\cite{lin2024rella} exemplifies this approach. Another line maintains user-specific adaptations. OPPU~\cite{tan2024democratizing} combines personal PEFT parameters with non-parametric knowledge from retrieval and profiles, while PER-PCS~\cite{tan2024personalized} composes shared personal parameter components to construct an adaptation suited to a target user's history.

Compared with full fine-tuning, these methods reduce the number of trainable parameters, but adaptation still depends on optimization and sufficient training evidence. Shared adapters must accommodate heterogeneous users, whereas user-specific adapters introduce storage, loading, and maintenance costs. Changes in user preferences can require renewed fitting or adapter recomposition. In contrast, \modelname leaves the base model fixed and represents personal evidence as a retrievable dictionary of activation directions. Updating this dictionary requires vector preparation rather than per-user gradient-based training.

\subsection{Personalized Reward Modeling}
Personalized reward modeling represents user preferences through a scoring function over candidate responses. Such scores can guide rejection sampling~\cite{liu2023statistical} or best-of-$N$ selection~\cite{gui2024bonbon}. Hydra~\cite{zhuang2024hydra}, for example, learns user-specific scoring adapters from positive and negative examples. Other approaches use an LLM as a personalized judge, conditioning its evaluation on user information rather than necessarily training a separate reward model~\cite{ryan2025synthesizeme}.

Learned reward models depend on the availability and coverage of preference feedback, while both learned and prompted judges must distinguish user-specific suitability across diverse outputs. In candidate-selection pipelines, quality is bounded by the available candidates, and increasing their number raises generation and evaluation costs. \modelname instead steers the generation process itself, without requiring explicit pairwise preference labels or multiple completed candidates for each inference query. This distinction concerns the illustrated inference pipeline rather than a claim that all uses of reward models require best-of-$N$ decoding.

\subsection{Inference-Time Activation Steering}
Activation steering modifies intermediate representations to influence a model's behavior without updating its parameters. Prior work studies additive activation interventions and contrastive direction construction~\cite{turner2023steering,rimsky2024steering,frandsen2022extracting}, while representation engineering explores the monitoring and control of high-level model representations~\cite{zou2023representation}. Persona- and style-oriented approaches further investigate steering directions associated with behavioral traits or personalized generation~\cite{chen2025persona,zhang2025personalized}.

Within this family, methods differ in how contrasts are constructed, where directions are extracted and injected, and whether the intervention varies across inputs. Figure~\ref{fig:intro} organizes the design of \modelname along these three axes. It contrasts contexts while holding a historical query and response fixed, preserves separate post-attention and MLP-output directions, and retrieves historical directions according to the current query. These choices distinguish its matched, fine-grained, instance-tailored formulation from response-contrast, whole-layer, or fixed user-level alternatives. The basic steering formulation is introduced next; the concrete construction and inference procedure are given in Section~\ref{sec:methodology}.

\section{Preliminary}
\label{sec:preliminary}

\subsection{Personalization Setting}
\label{sec:problem_formulation}
Let $\mathcal{M}_\theta$ denote a base language model with parameters $\theta$. For user $u$, let
\begin{equation}
    D_u=\{(X_i,Y_i)\}_{i=1}^{n_u}
\end{equation}
denote the historical interactions available before the current query $X$, where $X_i$ is a historical query and $Y_i$ is its recorded response. The output $Y$ may be natural-language text or a task-specific textual serialization of a structured action. The personalization objective is to generate a response aligned with the user's current, query-dependent preferences using the evidence in $D_u$.

The history provides observations rather than a complete specification of those preferences. Its relevance can vary across queries and over time. Under the temporal evaluation protocol, only information available before the evaluated interaction may be used for personalization. We write $D_u$ without an explicit time index for simplicity. Historical responses are available during preparation; the unknown response to the current query is not available for retrieval or direction construction.

\subsection{Personalization Interfaces}
The paradigms in Figure~\ref{fig:intro} differ in the interface through which user evidence influences generation. Let $\phi(X,D_u)$ retrieve or summarize the personal context relevant to $X$, and let $\Vert$ denote textual concatenation. Prompt-based personalization conditions the model on
\begin{equation}
    P_u(X)=X\Vert\phi(X,D_u),
    \qquad Y\sim\mathcal{M}_\theta(\cdot\mid P_u(X)).
    \label{eq:prompt_personalization}
\end{equation}
Actual inputs use the model's formatting template; the concatenation notation abstracts away those formatting details.

For user-specific parametric adaptation, personal evidence is encoded in a parameter change $\delta_u$:
\begin{equation}
    Y\sim\mathcal{M}_{\theta+\delta_u}(\cdot\mid P_u(X)).
    \label{eq:parametric_personalization}
\end{equation}
The prompt may be the query alone or may additionally contain retrieved context. Shared-adapter methods instead learn an adaptation across users and use conditioning information to distinguish their requests.

For reward-based candidate selection, let $\{Y^{(j)}\}_{j=1}^{N}$ be candidate responses generated for the current query and let $r_u(X,Y^{(j)})$ be a user-conditioned score. Best-of-$N$ selection takes the form
\begin{equation}
    Y^*=\underset{Y^{(j)},\,1\leq j\leq N}{\arg\max}
    \;r_u(X,Y^{(j)}).
    \label{eq:reward_personalization}
\end{equation}
The scorer may be learned from preference data or implemented by a prompted judge. This expression describes the candidate-selection setting, not every possible use of a reward model.

\subsection{Activation Steering}
\label{sec:steering}
Activation steering introduces personalization at an internal computational interface instead of requiring a parameter update. At a selected layer $L$, let $h_t^L\in\mathbb{R}^d$ denote the activation at token position $t$. A basic additive intervention is
\begin{equation}
    \widetilde{h}_t^L=h_t^L+\gamma\Delta^L,
    \label{eq:basic_steering}
\end{equation}
where $\Delta^L\in\mathbb{R}^d$ is a steering direction and $\gamma\in\mathbb{R}$ controls its strength. The modified activation is passed to the remaining computation while $\theta$ remains unchanged. The layer, interface, token positions, and intervention strength are design choices rather than universal constants.

A common construction uses a contrastive pair of complete inputs $Z^+$ and $Z^-$:
\begin{equation}
    \Delta^L=\overline{h}^{L}(Z^+)
             -\overline{h}^{L}(Z^-),
    \label{eq:generic_contrast}
\end{equation}
where $\overline{h}^{L}$ denotes an activation pooled over a specified token set. The meaning of this direction depends on what differs between the paired inputs and which activations are pooled. In a response contrast, the continuations differ; in the matched contrast used by \modelname, the historical query and response remain fixed while the conditioning contexts differ. Section~\ref{sec:methodology} specifies the context construction, response-token pooling, and two extraction interfaces.

The linear representation hypothesis motivates additive steering by suggesting that some high-level attributes are approximately encoded as directions in activation space~\cite{turner2023steering,rimsky2024steering}. Separately, a local approximation characterizes how a sufficiently small intervention propagates. Let $F_L$ map the activation at the selected interface to downstream logits, with the other inputs fixed. If $F_L$ is twice continuously differentiable with bounded second derivative in a neighborhood of $h$, then for a perturbation $\delta$ within that neighborhood,
\begin{equation}
    F_L(h+\delta)-F_L(h)
    =J_{F_L}(h)\delta+\mathcal{O}(\|\delta\|_2^2),
    \label{eq:local_steering}
\end{equation}
where $J_{F_L}(h)$ is the local Jacobian. This approximation does not establish global linearity, guarantee a monotonic behavioral change, or show that a contrast isolates user preference. Steering quality still depends on the evidence, extraction interface, intervention strength, and downstream nonlinear computation.

For personalization, a fixed user-level direction can be replaced by a query-conditioned direction $\Delta_u^L(X)$ obtained from historical evidence. \modelname instantiates this idea with a dictionary of paired, fine-grained directions and query-dependent retrieval. It also retains retrieved textual context in the generation prompt, so activation-space steering and prompt-based conditioning act as complementary information channels. The sequential dual-injection procedure is detailed in Section~\ref{sec:steering_foundation}.

\section{Methodology}
\label{sec:methodology}

\subsection{Overview}
\begin{figure*}[!ht]
    \centering
    \includegraphics[width=0.98\linewidth]{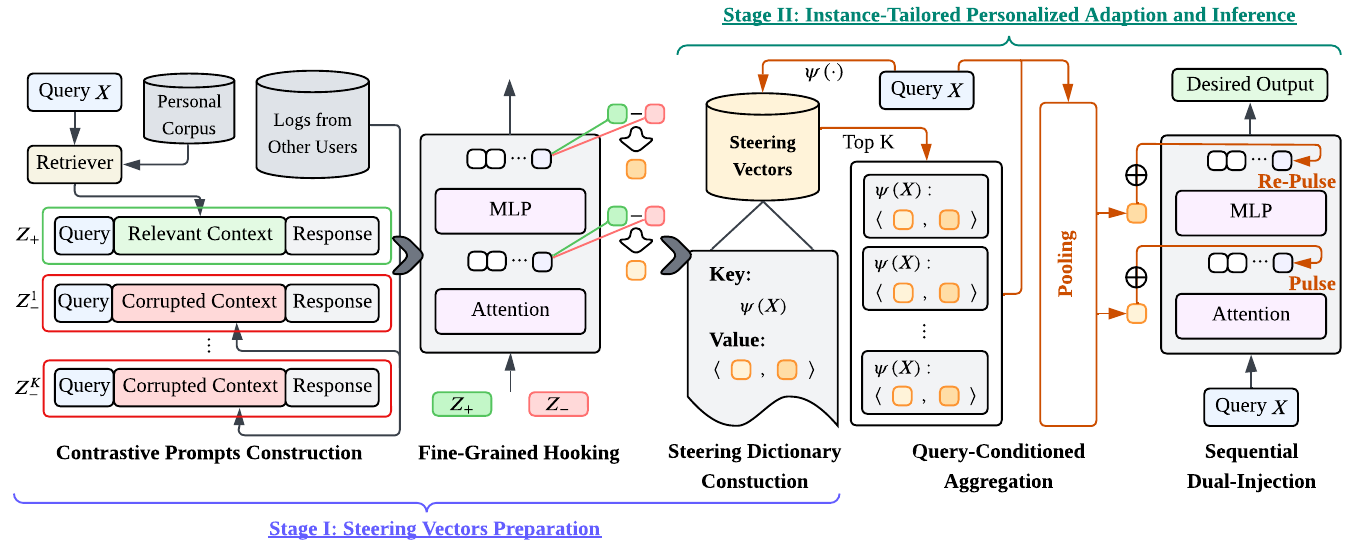}
    \caption{Overview of \modelname. 1) Steering Vectors Preparation. During this stage, we construct contrastive prompts from user logs, where relevant context retrieved from personal corpus is concatenated with target query to form the positive sample and irrelevant context sampled from other users is concatenated with target query to form the negative sample. For each positive sample, we generate $K$ negative samples. Each pair is then fed into the LLM in the teacher forcing mode, during which the response tokens' activations of attention and MLP blocks are hooked. We store the difference between two sample activations of a pair to serve as the steering vector, with the text of each pair being the key for indexing convenience. 
    2) Instance-Tailored Personalized Adaptation and Inference. During this stage, we sample from target user's steering vectors set to interfere model for personalized output. Concretely, we rank the similarity between query text and sample pair text to select the top-k steering vectors, which are then aggregated and injected into LLM for personalized adaptation.}
    \label{fig:framework}
    \vspace{-10pt}
\end{figure*}

Despite the increasing demand for personalized LLMs services, existing methods face critical problems in dynamic and data-sparse environments. Limitations of existing methods include structural interference, slow adaptation, and context window constraints, which hinder their applicability in real-world scenarios with fast-changing user interests and limited interaction data. In this paper, we propose \modelname, a novel inference-time personalization framework that treats personalization as a sample-level activation shift rather than a parametric update. \modelname operates through a two-stage pipeline designed for instance-aware personalization without model training:

\vspace{-3pt}

\begin{itemize}[leftmargin=*]
    \item \textbf{Offline Steering Vectors Preparation}: For each user, we construct a personalized dictionary of steering vectors from historical interactions. Each vector is derived by contrasting model activations under user-relevant and user-irrelevant contexts, using a \textbf{Fine-Grained Hooking} mechanism that separately extracts signals from attention and MLP layers.
    
    \item \textbf{Online Inference with Instance-Tailored Dual Steering}: During inference, for a target query, \modelname dynamically selects and aggregates the most relevant steering vectors from the user's dictionary based on semantic similarity. The aggregated vectors are then injected into the model's forward pass at a predefined layer via a \textbf{Sequential Dual-Injection} mechanism, enabling real-time, query-specific personalization. 
    This approach ensures that each query receives a tailored activation shift, offering instance-level adaptation. The position-matched injection offers reinforced interference with subtle signals from each component, under the constraints of geometric consistency and linear compatibility.
\end{itemize}

\vspace{-3pt}

These designs enable \modelname to adapt instantly with fine-grained signals to meet evolving user interests, with minimal memory and latency overhead, making it particularly suitable for dynamic and data-sparse environments where traditional parametric methods struggle.

\subsection{Steering Vectors Preparation}
Based on the time-splitting setting, we use old user logs for steering vectors preparation. 
\subsubsection{Contrastive Prompt Construction}
We construct contrastive prompt by varying the conditioning context while holding the historical query and target response fixed. For user $u$, let $(X_i,Y_i)$ denote a historical interaction, where $X_i$ is the query and $Y_i$ is its recorded response. A relevant context is retrieved from the user's history, excluding the current interaction, whereas a negative context is sampled from other users' histories:
\begin{align}
    C_i^+ &= \operatorname{Retrieve}
      \bigl(X_i,D_u\setminus\{(X_i,Y_i)\}\bigr),\\
    C_i^- &= \operatorname{Sample}(D_{\setminus u}).
\end{align}
Here, $D_{\setminus u}$ denotes the available histories of users other than $u$. The positive context supplies user-relevant evidence, while the negative context serves as a user-irrelevant contrast. Excluding the current interaction prevents its target response from being directly retrieved into the conditioning history. All context candidates must be restricted to the information available under the evaluation protocol.

Using the same formatting template, we construct
\begin{align}
    Z_i^+ &= \operatorname{Format}(X_i,C_i^+,Y_i),\\
    Z_i^- &= \operatorname{Format}(X_i,C_i^-,Y_i).
    \label{eq:contrastive_inputs}
\end{align}
The superscripts $+$ and $-$ refer to the contextual conditions, not to different responses: the same $X_i$ and $Y_i$ are supplied in both inputs. For structured outputs, $Y_i$ denotes the recorded response in its task-specific textual serialization, which is also kept identical across the pair.


This construction examines how relevant versus irrelevant conditioning changes the model's representation of a fixed response, rather than comparing independently generated continuations.

\subsubsection{Fine-Grained Activation Hooking}
\label{sec:fine_hook}

Consider a pre-normalized Transformer block at a selected layer $L$ for a complete input sequence $Z$. Let $r_t^L(Z)\in\mathbb{R}^d$ denote the incoming residual state at token $t$, and let $r_{\leq t}^L(Z)$ denote the corresponding causal prefix. The block computes
\begin{align}
    a_t^L(Z)
    &=\operatorname{Attn}^L\!\left(
      \operatorname{LN}_a^L(r_{\leq t}^L(Z))\right)_t,\\
    u_t^L(Z)&=r_t^L(Z)+a_t^L(Z),
    \label{eq:attn_hook}\\
    m_t^L(Z)&=g^L(u_t^L(Z)),\\
    v_t^L(Z)&=u_t^L(Z)+m_t^L(Z),
    \label{eq:mlp_hook}
\end{align}
where $g^L=\operatorname{MLP}^L\circ\operatorname{LN}_m^L$. These equations distinguish three interfaces: the post-attention residual $u_t^L$, the MLP output $m_t^L$ before residual addition, and the post-block residual $v_t^L$. \modelname extracts directions from $u_t^L$ and $m_t^L$, preserving the signal entering the MLP separately from the signal it contributes to the residual stream.

We process each complete input $Z_i^{\pm}$ with the frozen language model in teacher-forcing mode: the recorded response $Y_i$ is supplied in both passes rather than sampled from the model. The \textbf{fine-grained hooking} vectors are then captured from the \textbf{matched contrast} mode.

\begin{itemize} [leftmargin=10pt]
    \item \textbf{Matched Contrast.} Let $\mathcal{T}_i^+$ and $\mathcal{T}_i^-$ denote the nonempty position sets corresponding to $Y_i$ in the respective fully formatted and tokenized inputs. Only response-token activations are pooled; query, context, formatting-only, and padding positions are excluded. 

Our working hypothesis is that, for a fixed query and response, contrasting relevant and irrelevant contexts reveals an activation displacement associated with user-specific conditioning. Holding the query and response fixed removes differences in their content from the pair construction. Interpreting the remaining contrast as a preference-related displacement additionally requires other contextual effects to be balanced in expectation.

For a fixed user, layer, and interface, consider the idealized decomposition
\begin{equation}
    \overline{h}_{p,i}^{L,\pm}
    =\mu_p^L(X_i,Y_i)+\rho_{p,i}^{L,\pm}
      +\varepsilon_{p,i}^{L,\pm},
\end{equation}
where $\mu_p^L$ is a shared query--response component, $\rho_{p,i}^{L,\pm}$ is the preference-related contribution under each context, and $\varepsilon_{p,i}^{L,\pm}$ collects remaining contextual effects. If
\begin{equation}
    \mathbb{E}\!\left[
      \varepsilon_{p,i}^{L,+}-\varepsilon_{p,i}^{L,-}
      \mid X_i,Y_i\right]=0,
\end{equation}
then the contrastive estimator satisfies
\begin{equation}
    \mathbb{E}[\Delta_{p,i}^L\mid X_i,Y_i]
    =\mathbb{E}\!\left[
      \rho_{p,i}^{L,+}-\rho_{p,i}^{L,-}
      \mid X_i,Y_i\right].
\end{equation}
The expectation is over context-pair construction for the fixed interaction. This is a conditional identification statement, not a guarantee of the construction procedure: context topic, length, factual content, and induced positional shifts may contribute systematically to the contrast. Accordingly, the method uses empirical steering directions without claiming that every direction isolates an unbiased preference effect.

\item \textbf{Fine-Grained Hooking.}  The term \emph{fine-grained} refers to retaining two within-block contrastive signals rather than collapsing them into a single post-block difference. Attention steering modulates the query-key-value interactions (capturing stylistic focus patterns), while MLP steering adjusts the feed-forward transformations (capturing lexical and semantic preferences). These operate on distinct representational subspaces. We therefore capture these complementary signals.


Concretely, for $p\in\{attn,mlp\}$, define
\begin{equation}
    \overline{h}_{p,i}^{L,\pm}
    =\frac{1}{|\mathcal{T}_i^{\pm}|}
      \sum_{t\in\mathcal{T}_i^{\pm}}h_{p,t}^L(Z_i^{\pm}),
\end{equation}
where $h_{attn,t}^L=u_t^L$ and $h_{mlp,t}^L=m_t^L$. The paired steering directions are
\begin{align}
    \Delta_{attn,i}^L
    &=\overline{h}_{attn,i}^{L,+}
      -\overline{h}_{attn,i}^{L,-},
    \label{eq:delta_attn}\\
    \Delta_{mlp,i}^L
    &=\overline{h}_{mlp,i}^{L,+}
      -\overline{h}_{mlp,i}^{L,-}.
    \label{eq:delta_mlp}
\end{align}
The two response representations are averaged separately before subtraction. 

\end{itemize}

\subsubsection{Steering Dictionary Construction}

To support query-conditioned retrieval, each record is associated with the representation
\begin{equation}
    e_i=Z_i^+,
    \qquad \mathbf{k}_i=\psi(e_i),
    \label{eq:encoding}
\end{equation}
where $\psi$ is the retrieval encoder, distinct from the frozen language model used for activation extraction. The retrieval text is the positive-context sequence, containing $X_i$, $C_i^+$, and the historical response $Y_i$. The negative context is used to estimate the activation contrast but is not included in this retrieval key. The historical response is available during preparation; the target response to the current inference query is never used for retrieval. At layer $L$, the user-specific dictionary is
\begin{equation}
    \mathcal{S}_u=
    \left\{\left(\mathbf{k}_i,
      \Delta_{attn,i}^L,\Delta_{mlp,i}^L\right)
    \right\}_{i=1}^{N_u}.
\end{equation}
Thus, each retained contrastive record provides both a semantic retrieval key and a pair of activation-space directions. This separation allows the historical evidence to be indexed in the encoder's representation space while its behavioral effect is expressed in the base model's hidden space. For clarity, the subsequent formulation uses one selected layer; the same construction applies separately to each layer when multiple layers are used.

\subsection{Input-Aware Personalized Inference}

\subsubsection{Query-Conditioned Retrieval and Aggregation}

Given a query $X$, \modelname selects relevant entries from $\mathcal{S}_u$ rather than applying a fixed average over the user's entire history. Let $\bar X$ denote the task-specific retrieval text extracted from $X$. We use a Contriever encoder with mean pooling over non-padding tokens and score each entry by the unnormalized inner product
\begin{equation}
    s_i(X)=\psi(\bar X)^\top\mathbf{k}_i.
\end{equation}
For a nonempty dictionary and a positive retrieval budget $K$, let $K_u=\min(K,N_u)$. The selected index set is
\begin{equation}
    \mathcal{I}_u(X)=
    \underset{i\in\{1,\ldots,N_u\}}{\operatorname{TopK}}
    \bigl(s_i(X),K_u\bigr),
\end{equation}
where $\operatorname{TopK}$ returns the indices of the $K_u$ highest scores. The selected directions are aggregated independently at the two interfaces:
\begin{equation}
    \mathbf{s}_p^L(X)=\frac{1}{K_u}
      \sum_{i\in\mathcal{I}_u(X)}\Delta_{p,i}^L,
    \quad p\in\{attn,mlp\}.
    \label{eq:query_aggregation}
\end{equation}
The generation prompt additionally includes retrieved textual history:
\begin{equation}
    P_u(X)=\operatorname{Format}\bigl(X,H_u(X;B)\bigr),
\end{equation}
where $H_u(X;B)$ denotes history retrieved with a prompt budget $B$. The budgets $B$ and $K$ govern different information channels: the former determines explicit contextual evidence, whereas the latter determines activation-space steering. Their retrieved records need not coincide. This distinction makes the role of steering explicit even when it is combined with retrieval-augmented prompting.

\subsubsection{Sequential Dual-Injection}
\label{sec:steering_foundation}
Afterwards, we perform a sequential dual-injection for inference-time enhancement. Overall, this position-matched injection ensures:
\begin{itemize}[leftmargin=10pt]
    \item \textbf{Geometric consistency}: The intervention occurs at the same representational locus where the steering signal was derived.
    \item \textbf{Linear compatibility}: Per the linear representation hypothesis, additive vectors in the same latent space propagate without nonlinear distortion through subsequent layers.
\end{itemize}

Concretely, at layer $L$, each aggregated direction is applied at the same computational interface from which it was extracted. For a token position $t$ receiving an intervention, sequential dual-injection computes
\begin{align}
    \widetilde{u}_t^L
    &=u_t^L+\alpha\mathbf{s}_{attn}^L(X),
    \label{eq:dual_attn}\\
    \widetilde{m}_t^L
    &=g^L(\widetilde{u}_t^L)
      +\beta\mathbf{s}_{mlp}^L(X),\\
    \widetilde{v}_t^L
    &=\widetilde{u}_t^L+\widetilde{m}_t^L.
    \label{eq:dual_mlp}
\end{align}
The modified block output is then passed to the remaining layers and the language-model head. The coefficients $\alpha$ and $\beta$ independently regulate the attention-side and MLP-side interventions. Setting both coefficients to zero recovers the unsteered forward computation for the same generation prompt; setting either coefficient to zero yields the corresponding single-site intervention.

The order of these operations is essential. The attention-side perturbation first changes the residual entering the MLP, so the MLP is evaluated on $\widetilde{u}_t^L$, not on the original $u_t^L$. The MLP-side direction is then added to this recomputed output before residual addition. The resulting operation is therefore not, in general, equivalent to adding $\alpha\mathbf{s}_{attn}^L+\beta\mathbf{s}_{mlp}^L$ to the unmodified block output.

In the cached dual-site configuration, these interventions are applied during both prompt processing and autoregressive decoding. This inference-time scope is distinct from the response-token ranges used to estimate the directions. Matching extraction and injection interfaces establishes where each vector acts; it does not require identical token ranges between preparation and inference.

\section{Experiment}
\subsection{Setup}
\subsubsection{Datasets}
To evaluate \modelname under different scenarios, we utilize datasets including short personalized content generation (headline generation\footnote{https://lamp-benchmark.github.io/}), long personalized content generation (abstract writing\footnote{https://longlamp-benchmark.github.io/download}), and personal web function calling (PersonalWAB\footnote{https://github.com/HongruCai/PersonalWAB/tree/main/PersonalWAB}) for evaluation. Each of them contains user id, corresponding profile, and interaction histories. 

\begin{table}[h!]
\centering
\caption{Statistics of Used Datasets.}
\label{tab:sta}
\resizebox{0.99\linewidth}{!}{
\begin{tabular}{c|cccc}
\hline

Datasets                                                                            & \#Users & \makecell[c]{Avg. \\Profiles} & \makecell[c]{Avg. Text\\Length \\(per user)} & \makecell[c]{Avg. Text\\Length \\(per sample)} \\ \hline
\makecell[c]{Headline\\Generation}                                                         & 200     & 68.26         & 2,257                                                                  & 33                                                                      \\\hline
\makecell[c]{Abstract\\Writing}                                                            & 200     & 117.91        & 18,369                                                                 & 156                                                                     \\\hline
\begin{tabular}[c]{@{}c@{}}PersonalWAB\\ (product)\end{tabular}     & 200     & 60.72         & 20,348                                                                 & 335                                                                     \\\hline
\begin{tabular}[c]{@{}c@{}}PersonalWAB\\ (interaction)\end{tabular} & 200     & 12.2          & 656                                                                   & 54                                                                      \\ \hline
\end{tabular}
}
\end{table}
\begin{itemize}[leftmargin=10pt]
    \item \textbf{News-Headline Generation}. This dataset~\cite{misra2022news} evaluates the ability of a large language model to capture the stylistic patterns of an author, querying the LLM to generate a headline given the input news article. 
    \item \textbf{Abstract Writing}. This dataset evaluates LLMs' ability to distill complex ideas and generate accurate, concise, and coherent output over the span of multiple paragraphs on domain-specific tasks. The input of the model is the title of the paper along with some keywords to guide the content. The expected output is an abstract conditioned on the title and keywords in the user's style\cite{kumar2024longlamp}.
    \item  \textbf{PersonalWAB}. This dataset integrates personalized instruction comprehension and action execution, where LLMs
    must infer personalized user requirements and preferences to determine which Web function to call and formulate the corresponding function parameters. Subsequently, the results of these function calls are returned to users\cite{cai2025large}.
\end{itemize}

We sample 200 users from the original dataset, based on the time-split setting, earlier question-answer~(QA) pairs form the train set and the latter pairs form the test set. The statistics of used datasets are summarized in Table~\ref{tab:sta}.

\vspace{-10pt}
\subsubsection{Baselines}
We evaluate \modelname against: 1) Prompt-based method (In Context Learning) and 2) Parametric adaptation methods including One-Peft-All-User\cite{lin2024rella} and One-Peft-Per-User \cite{tan2024democratizing,tan2024personalized}. 3) Personalized reward models\cite{ryan2025synthesizeme},
3) Steering-based baseline \cite{zou2023representation,chen2025persona,zhang2025personalized}.

\vspace{-10pt}
\subsubsection{Metric}
We use Rouge Values \cite{lin2004rouge} and METEOR \cite{banerjee2005meteor} to measure the quality of content generation and use Overall Accuracy (ACC) to measure the quality of web function calling.

\begin{table*}[!t]
\centering
\caption{Main results on content generation and personal web function calling. All baselines are done with n-shot demonstrations, consistent with the ICL setting. (The symbol * indicates statistically significant improvement with p-value $\le 0.001$. Hyperparameters sensitivity is discussed in Appendix~\ref{sec:hyper}.)}
\label{tab:main}
\resizebox{0.99\linewidth}{!}{
\begin{tabular}{cccc|ccc|ccc|c}
\hline
\multicolumn{4}{c|}{Datasets}                                                                                                                                         & \multicolumn{3}{c|}{Headline Generation} & \multicolumn{3}{c|}{Abstract Writing} & PersonalWAB \\ \hline
\multicolumn{4}{c|}{Methods}                                                                                                                                          & Rouge-1        & Rouge-L    & METEOR  & Rouge-1                 & Rouge-L     & METEOR     & ACC   \\ \hline
\multicolumn{1}{c|}{Direct}                               & \multicolumn{3}{c|}{ZeroShot}                                                                           & 0.1432         & 0.1297 &0.1002        & 0.3503                  & 0.2024      &0.1999          & 0.6017\\ \hline
\multicolumn{1}{c|}{\multirow{4}{*}{\makecell[c]{In Context\\ Learning}}} & \multicolumn{3}{c|}{k=1}                                                                                & 0.1389         & 0.1287 &0.0994        & 0.3648                  & 0.2093      &0.2298         &0.6767  \\
\multicolumn{1}{c|}{}                                       & \multicolumn{3}{c|}{k=3}                                                                                & 0.1643         & 0.1515 &0.1257        & 0.3701                  & 0.2104      & 0.2443      &0.8415     \\
\multicolumn{1}{c|}{}                                       & \multicolumn{3}{c|}{k=5}                                                                                & 0.1665         & 0.1506 & 0.1267        & 0.3771                  & 0.2153       & 0.2517        &0.8330  \\
\multicolumn{1}{c|}{}                                   & \multicolumn{3}{c|}{k=10}          &       0.1630         & 0.1488     & 0.1271         &         0.3768                &     0.2127     &0.2559         &0.7966     \\ \hline
\multicolumn{1}{c|}{\multirow{3}{*}{\makecell[c]{Parametric Adaptation}}}  & \multicolumn{3}{c|}{Rella}  &0.1635&	0.1463	& 0.1263&		0.3818&	0.2141&	0.2562 &	0.8330 \\
\multicolumn{1}{c|}{}   & \multicolumn{3}{c|}{OPPU}                                                                               & \underline{0.1779}         & \underline{0.1613} & \underline{0.1348}        &         0.3848	&0.2146         & \underline{0.2580} & 0.8394    \\
\multicolumn{1}{c|}{}                                       & \multicolumn{3}{c|}{PER-PCS}                                                                            & 0.1763         & 0.1577 & 0.1255        &         0.3602                &  0.2169    &0.2446           &0.8402       \\ \hline
\multicolumn{1}{c|}{Reward Models} & \multicolumn{3}{c|}{SynthesizeMe} & 0.1672&	0.1490&	0.1288&	0.3906&	0.2156&	0.2565&	0.8275\\ \hline
\multicolumn{1}{c|}{\multirow{7}{*}{\makecell[c]{Steering}}}  & \multicolumn{3}{c|}{REPE} &0.1685&	0.1531&	0.1325&	0.3783&	0.2134&	0.2470&	\underline{0.8501}\\
\multicolumn{1}{c|}{}   & \multicolumn{3}{c|}{PersonaVector (Response)}  &0.1714	&0.1540	&0.1339	& \underline{0.3954}	&\underline{0.2291}	&0.2567	&0.8116 \\ 
\multicolumn{1}{c|}{}   & \multicolumn{3}{c|}{PersonaVector (All)} & 0.1700	&0.1536	&0.1306	&0.3825	&0.2169	&0.2512	&0.8308\\ 
\multicolumn{1}{c|}{}   & \multicolumn{3}{c|}{PersonaVector (Prompt)} & 0.1640	&0.1481	&0.1267	&0.3834	&0.2148	&0.2494	&0.8458 \\ 
\multicolumn{1}{c|}{}   &\multicolumn{3}{c|}{StyleVector (Mean)}  &0.1694&	0.1540	&	0.1345&	0.3790&	0.2178&	 0.2547&	0.8137 \\
\multicolumn{1}{c|}{}   & \multicolumn{3}{c|}{StyleVector (Logistic)}  &0.1712&	0.1539	&	0.1342&	0.3831&	0.2136&	 0.2505&	0.8436 \\ \cline{2-11}
\multicolumn{1}{c|}{} &\multicolumn{3}{c|}{\modelname} & \textbf{0.1816*}                   & \textbf{0.1666*}     & \textbf{0.1473*}              &       \textbf{0.3990*}                  &        \textbf{0.2306*}     &\textbf{0.2609*}         &\textbf{0.8566*}  \\ \hline
\end{tabular}
}
\end{table*}

\begin{table*}[!t]
\caption{Impact of hooking strategy. (The symbol * indicates statistically significant improvement with p-value $\le 0.001$.)}
\label{tab:finegrained}
\resizebox{0.99\linewidth}{!}{
\begin{tabular}{cccc|ccc|ccc|c}
\hline
\multicolumn{4}{c|}{Datasets}                                                                                                                                         & \multicolumn{3}{c|}{Headline Generation} & \multicolumn{3}{c|}{Abstract Writing} & PersonalWAB \\ \hline
\multicolumn{4}{c|}{Methods}                                                                                                                                          & Rouge-1        & Rouge-L & METEOR       & Rouge-1                 & Rouge-L     & METEOR        & ACC   \\ \hline
\multicolumn{1}{c|}{\multirow{4}{*}{\makecell[c]{\modelname}}}                                       & \multicolumn{2}{c|}{\multirow{4}{*}{\makecell[c]{Fine-Grained \\Hooking}}} & Post-Attention      & 0.1752                   & 0.1590 &0.1384                    & 0.3938                  & 0.2271       &0.2556    &\textbf{0.8588}      \\
\multicolumn{1}{c|}{}                                       & \multicolumn{2}{c|}{}                                     & Post-MLP                   & 0.1704                   & 0.1544        &0.1301           & 0.3826                  & 0.2133     &0.2503      & 0.8501    \\ 
\multicolumn{1}{c|}{}                                    & \multicolumn{2}{c|}{}                                     & Post-Block                 & 0.1737                   & 0.1575           &0.1369        & 0.3946                  & 0.2277     &0.2556        &0.8458    \\ \cline{4-11}
\multicolumn{1}{c|}{}                                    & \multicolumn{2}{c|}{}                                     & Hybrid Strategy            & \textbf{0.1816*}                   & \textbf{0.1666*}        &\textbf{0.1473*}           &       \textbf{0.3990*}                  &        \textbf{0.2306*}      &\textbf{0.2609*}        &0.8566  \\ \hline
\end{tabular}
}
\end{table*}

\subsection{Research Questions}
\begin{itemize}[leftmargin=10pt]
    \item \textbf{(RQ1)} Does \modelname outperform the competing baselines?
    \item \textbf{(RQ2)} Does fine-grained hooking bring performance gain?
    \item  \textbf{(RQ3)} How is the robustness of \modelname against heterogeneous/fast changing distribution compared to other methods?
    \item \textbf{(RQ4)} How is the data efficiency of \modelname compared to other methods? Particularly, under high data sparsity?
    \item \textbf{(RQ5)} How is the overhead brought by \modelname? Is \modelname real-world feasible?
    \item \textbf{(RQ6)} How is the hyperparameters sensitivity of \modelname?
\end{itemize}

\vspace{-20pt}
\subsection{Main Results (RQ1-RQ2)}
The main results of the experiments are included in Table~\ref{tab:main}.


\minisection{\textbf{(RQ1)}} From the results, we can see that \modelname outperforms all other baselines. Concretely, parametric methods show better performances than the prompt-based methods, showcasing the benefits of personalization in parameters to better digest user patterns. Furthermore, \modelname outperforms traditional personalized methods, justifying the superiority of the steering-based methodology. 

\minisection{\textbf{(RQ2)}} 
To evaluate the efficacy of steering vectors beyond layer-wise granularity, we investigate different hooking strategies:

\begin{itemize}[leftmargin=10pt]
    \item \textbf{Post-Attention.} The direction is extracted from and injected into the post-attention residual $u^L$, after the attention skip addition and before the MLP normalization.
    \item \textbf{Post-MLP.} The direction is extracted from and injected into the raw MLP sub-layer output $m^L$, before its addition to the residual stream.
    \item \textbf{Post-Block.} The direction is extracted from and injected into the post-block residual $v^L=u^L+m^L$.
    \item \textbf{Hybrid Strategy.} Position-matched directions are injected sequentially at $u^L$ and $m^L$, following Equations~\eqref{eq:dual_attn}--\eqref{eq:dual_mlp}.
\end{itemize}


Table~\ref{tab:finegrained} reports the performance of the hooking variants. The Hybrid Strategy improves content-generation metrics by $0.36\%$--$1.64\%$ over the strongest single-site or post-block alternative ($p\le 0.001$). 


\subsection{Ablation Studies}
\subsubsection{(RQ3) Robustness to Drifting Distribution} In this section, we study the robustness of different methods against drifting distributions.  Particularly, we use the realistic temporal drift setting, where samples are split based on the timestamps and only earlier logs are available. We then test models' performances on different time ranges. The results of different methods are summarized in 
Table~\ref{tab:temporal}. 



\begin{table*}[!t]
\caption{Results of different methods against heterogeneous distribution among temporal dimension.}
\label{tab:temporal}
\resizebox{0.99\linewidth}{!}{
\begin{tabular}{cc|ccc|ccc|ccc}
\hline
\multicolumn{2}{c|}{(Headline Generation)}                                        & \multicolumn{3}{c|}{Time-Range 1} & \multicolumn{3}{c|}{Time-Range 2} & \multicolumn{3}{c}{Time-Range 3} \\ \hline
\multicolumn{2}{c|}{Methods}                                         & Rouge-1           & Rouge-L     & METEOR      & Rouge-1           & Rouge-L      & METEOR     & Rouge-1           & Rouge-L     & METEOR     \\ \hline 
\multicolumn{2}{c|}{In Context Learning}                                             & 0.1620&	0.1362&	0.1157	&0.1291	&0.1109	&0.0823	&0.1491	&0.1368	&0.1216    \\ \hline
\multicolumn{1}{c|}{\multirow{1}{*}{Parametric adaptation}} & OPPU     & 0.1751 &	0.1544&	0.1245&	0.1496&	0.1339&	0.1121&	0.1560&	0.1397&	0.1425     \\ \hline
\multicolumn{1}{c|}{\multirow{1}{*}{Personalized Reward Models}} & SynthesizeMe     & 0.1583 &	0.1380	&0.1109	&0.1497	&0.1355	&0.0974	&0.1551	&0.1336	&0.1295     \\\hline
\multicolumn{1}{c|}{\multirow{4}{*}{Steering}}        & REPE   & 0.1659	&0.1517	&0.1219	&0.1435	&0.1276	&0.0929	&0.1657	&0.1474	&0.1325       \\
\multicolumn{1}{c|}{}          & PersonaVector   & 0.1709 &	0.1536&	0.1220&	0.1475&	0.1363&	0.0928&	0.1642&	0.1403&	0.1341       \\
\multicolumn{1}{c|}{} & StyleVector   & 0.1556	&0.1444	&0.1118	&0.1508	&0.1350	&0.0961	&0.1602	&0.1359	&0.1272 \\\cline{2-11} 
\multicolumn{1}{c|}{}                         & \modelname & 0.1787	&0.1554	&0.1251	&0.1531	&0.1411	&0.1219	&0.1782	&0.1424	&0.1448       \\ \hline
\end{tabular}
}
\vspace{-10pt}
\end{table*}

\begin{figure}[!h]
    \centering
    \includegraphics[width=0.95\linewidth]{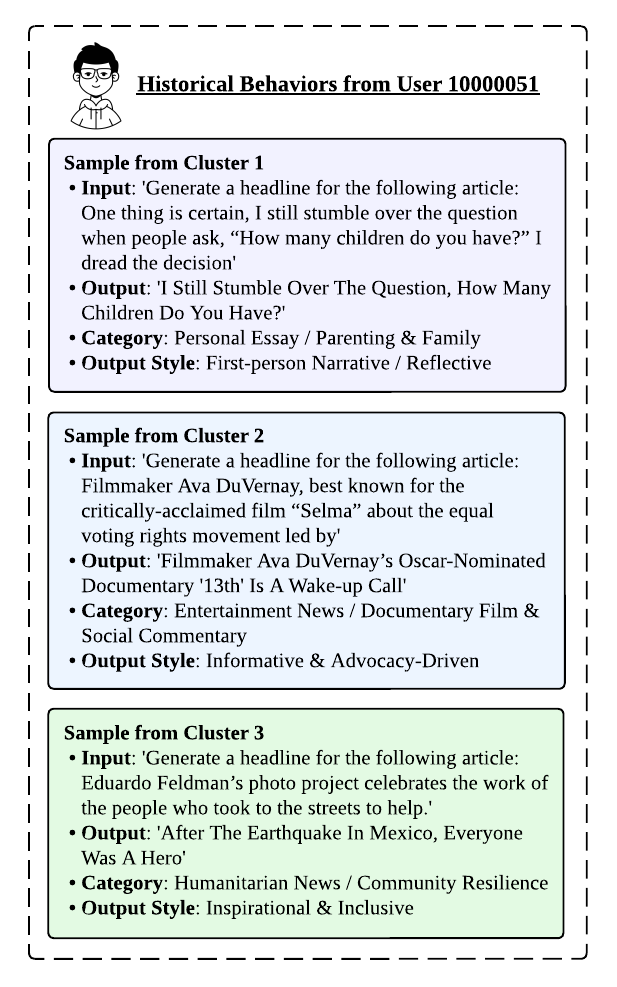}
    \caption{Illustration of the heterogeneous data.}
    \vspace{-20pt}
    \label{fig:case}
\end{figure}

From the results, we can see that \modelname outperforms all methods in each time-range, showing \modelname' strong superiority in realistic temporal drift setting. Since the parametric methods cannot quickly realign to rapidly changing user preferences due to their lock to sluggish and offline training loops. While \modelname offers inference-time  and instance-tailored adaptation, and therefore it offers flexibility and robustness to heterogeneous distributions. This feature makes \modelname distinctive in scenario where user patterns change fast, which is a commonly-seen and important problem in LLMs personalization.

To further illustrate the discrepancy existing in the drifting distributions, we select different samples of the same user $10000051$ as illustrated in Figure~\ref{fig:case}. These illustrations demonstrate the heterogeneous content themes, audience engagement strategies, and emotional tones existing in Question-Answer pairs of the same user.

\vspace{-5pt}

\subsubsection{(RQ4) Robustness to Data Scarcity and Data Efficiency Discussion}
\begin{figure}[h]
    \centering
    \vspace{-8pt}
    \includegraphics[width=0.99\linewidth]{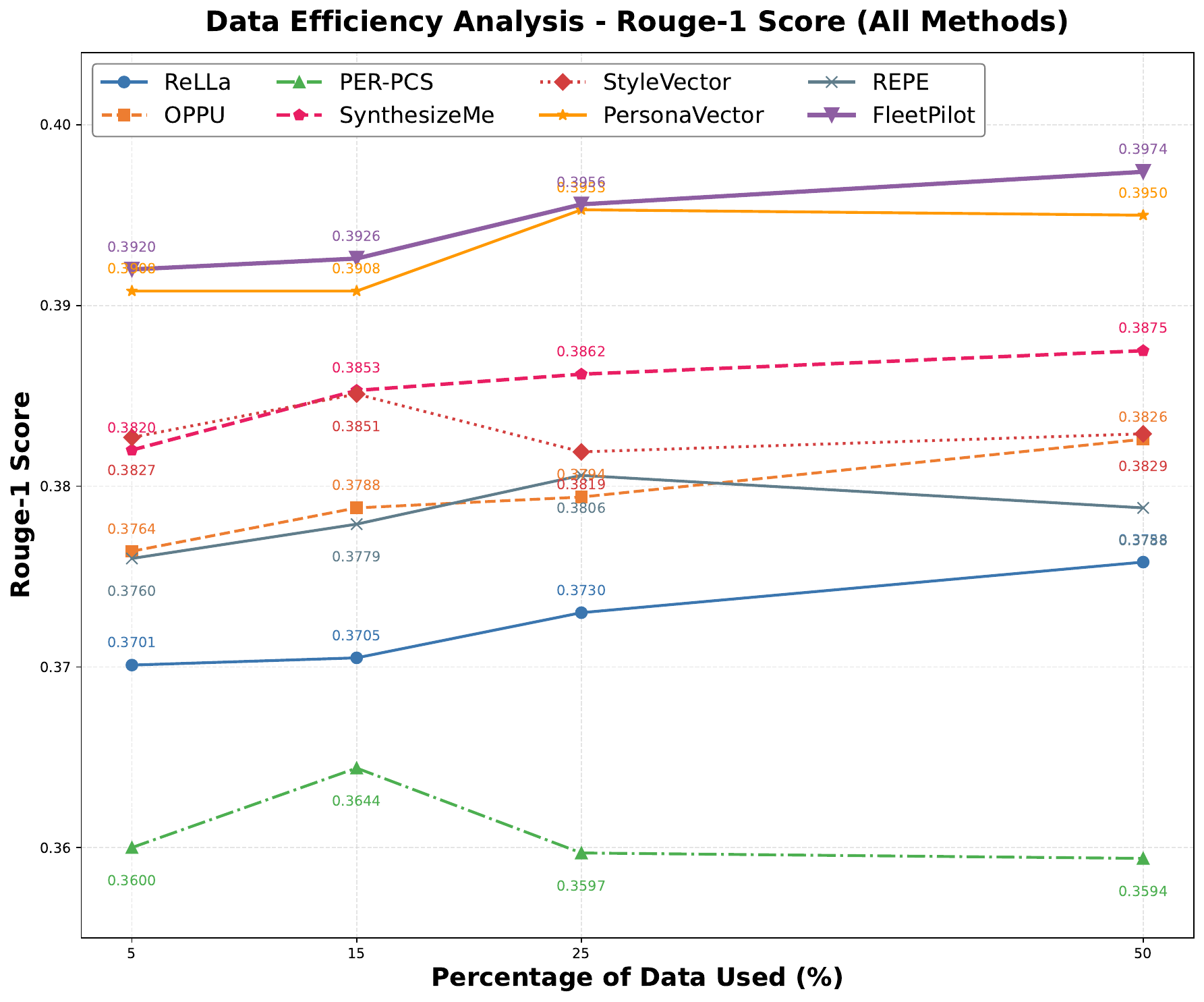}
    \vspace{-18pt}
    \caption{Data efficiency analysis of different methods.}
    \vspace{-12pt}
    \label{fig:df}
\end{figure}

Despite the robustness of heterogeneous distribution, data sparsity is also an important problem in LLMs personalization, where only a small amount of data is available to conduct effective personalization. Therefore, we study the data efficiency of different methods in this section. Concretely, we use different amounts of personalized data to train models/serve as steering materials. The results are summarized in Table~\ref{tab:df} and visualized in Figure~\ref{fig:df}.

\begin{table*}[!t]
\caption{Data efficiency analysis of different methods (Abstract Writing).}
\label{tab:df}
\resizebox{0.99\linewidth}{!}{
\begin{tabular}{cc|ccc|ccc|ccc|ccc}
\hline
\multicolumn{2}{c|}{\#Used Data}                                    & \multicolumn{3}{c|}{5\% (Avg. 5.82)}                        & \multicolumn{3}{c|}{15\% (Avg. 17.46)}                      & \multicolumn{3}{c|}{25\% (Avg. 29.10)}                      & \multicolumn{3}{c}{50\% (Avg. 58.72)}                      \\ \hline
\multicolumn{2}{c|}{Methods}                                        & \multicolumn{1}{c}{Rouge-1} & Rouge-L & METEOR & \multicolumn{1}{c}{Rouge-1} & \multicolumn{1}{c}{Rouge-L} & METEOR & \multicolumn{1}{c}{Rouge-1} & \multicolumn{1}{c}{Rouge-L} & METEOR& \multicolumn{1}{c}{Rouge-1} & \multicolumn{1}{c}{Rouge-L} & METEOR\\ \hline
\multirow{3}{*}{\makecell{Parametric\\ adaptation}} & ReLLa                           & 0.3701                       & 0.2036        &    0.2382          & 0.3705                       & 0.2134         &        0.2441       & 0.3730                      & 0.2137           &    0.2437          & 0.3758   & 0.2185   &  0.2508\\
 & OPPU                           & 0.3764                       & 0.2000        &0.2488                 & 0.3788                       & 0.2033         &0.2515                & 0.3794                       & 0.2132           &0.2523               & 0.3834   & 0.2200   & 0.2571\\
                                    & PER-PCS                        & 0.3600                       & 0.2154       &0.2445                  & 0.3644                       & 0.2158          &0.2491               & 0.3597                       & 0.2150       &0.2441                  & 0.3594                       & 0.2152              &0.2442          \\  \hline
\makecell{Personalized\\ Reward Models} & SynthesizeMe       & 0.3820                       & 0.2091        & 0.2564                 & 0.3853                       & 0.2107         & 0.2584               & 0.3862                       & 0.2109         & 0.2585              & 0.3875   & 0.2123   & 0.2575          \\ \hline
\multirow{4}{*}{Steering}            & StyleVector                           & 0.3827                       & 0.2151          &0.2499             & 0.3851                       & 0.2158         &0.2518                & 0.3819                       & 0.2147             &0.2488    & 0.3829 &0.2130 &0.2507 \\\cline{2-14} 
                                    & PersonaVector                         & 0.3908                       & 0.2256        & 0.2516                 & 0.3908                       & 0.2277         & 0.2543               & 0.3953                       & 0.2268         & 0.2544              & 0.3950   & 0.2288   & 0.2580          \\\cline{2-14}
                                    & REPE                                  & 0.3760                       & 0.2100        & 0.2446                 & 0.3779                       & 0.2103         & 0.2440               & 0.3806                       & 0.2133         & 0.2466              & 0.3788   & 0.2143   & 0.2495          \\\cline{2-14}
                                    & \multicolumn{1}{c|}{\modelname} & 0.3920 & 0.2236 & 0.2542 & 0.3926 & 0.2247 & 0.2561 & 0.3956 & 0.2257 & 0.2563 & 0.3974 & 0.2270 & 0.2594 \\ \hline
\end{tabular}
}
\end{table*}

The results show that \modelname remains effective with limited personalization histories, achieving the highest ROUGE-1 across all four data fractions. Particularly, under extreme data sparsity (<5 shots), steering presents strong superiority over the parametric methods.
The increasing trend indicates that \modelname preserves much of its performance under substantial
history reduction while still benefiting from additional evidence.

A plausible explanation for this effectiveness under sparse
histories is that activation steering reuses behavioral structure
already encoded in the pretrained model through a constrained
intervention, rather than estimating a parameter update from
limited user-specific data.
For \modelname, matched query--response contrasts may reduce
variation unrelated to contextual conditioning, while
response-token averaging may stabilize the extracted signals.
Query-conditioned retrieval further selects potentially relevant
historical directions instead of merging all available records
into a single user-level representation.
Together, these mechanisms offer a plausible account of its
effective use of limited histories.

\vspace{-10pt}
\subsubsection{(RQ5) Overhead Discussion and Deployment Analysis}
We sample 200 datapoints from each dataset and conduct the evaluation. The results are displayed in Table~\ref{tab:time}.

\begin{table}[h]
\centering
\caption{Latency Analysis.}
\label{tab:time}
\resizebox{0.99\linewidth}{!}{
\begin{tabular}{c|cc}
\toprule
\multirow{2}{*}{Datasets (200 samples)} & \multicolumn{2}{c}{Headline Generation}        \\ 
                                        & Direct Inference & + \modelname \\ \hline
Latency (s)                             & 63.23            & 72.49                       \\ \hline
&&  \\\\[-20pt]\hline
\multirow{2}{*}{Datasets (200 samples)} & \multicolumn{2}{c}{Abstract Writing}           \\ 
                                        & Direct Inference & + \modelname \\ \hline
Latency (s)                             & 670.34           & 712.36                      \\ \hline
&&  \\\\[-20pt] \hline 
\multirow{2}{*}{Datasets (200 samples)} & \multicolumn{2}{c}{PersonalWAB}                \\ 
                                        & Direct Inference & + \modelname \\ \hline
Latency (s)                             & 1161.88          & 1343.89                     \\ \bottomrule
\end{tabular}
}
\vspace{-10pt}
\end{table}

The inference overhead for \modelname is primarily attributed to retrieving the relevant steering vectors from the set, which is the most time-consuming step, and subsequently inserting the aggregated steering signal into the LLM's forward pass. While latency varies with the steering vector size per user and the query's token count, Table~\ref{tab:time} displays the results of time latency analysis brought by \modelname, which is conducted on a single NVIDIA H100 device. Results demonstrate that \modelname incurs only a light overhead during inference.  This efficiency offers a crucial advantage by circumventing the huge upload burden typically associated with transferring distinct personalized weights. 

We further compare the per-user storage overhead of \modelname with that of personalized parametric adaptation methods. Let $d$ denote the hidden dimension and $N$ denote the number of steering vectors retained for each user. A steering vector has $d$ elements, resulting in a storage complexity of $\mathcal{O}(d)$. In our implementation, $d=4096$ and each element is stored in \texttt{bfloat16} format, which occupies two bytes. Thus, each steering vector requires
\begin{equation}
    4096 \times 2 = 8192 \text{ bytes},
\end{equation}
or approximately $10$~KB on disk after accounting for serialization overhead. Since fewer than 100 steering vectors are retained for each user, the total per-user storage remains below $1$~MB. The resulting per-user storage complexity of \modelname is therefore
\begin{equation}
    \mathcal{O}(Nd).
\end{equation}

By contrast, personalized parametric adaptation methods must store a separate set of LoRA weights for each user. Let $r$ denote the LoRA rank and $L$ denote the number of adapted Transformer layers. The storage complexity of a personalized LoRA checkpoint is
\begin{equation}
    \mathcal{O}(rdL),
\end{equation}
up to constant factors determined by the number and dimensions of the adapted projection matrices. In our experiments, where $r=8$ and $L=32$, each user-specific LoRA weight dictionary occupies approximately $18$~MB on disk.

Because $N \ll rL$ under our configuration, \modelname requires substantially less per-user storage than personalized LoRA methods: less than $1$~MB compared with approximately $18$~MB per user. This difference becomes increasingly consequential as the number of users grows. Moreover, updating a user profile in \modelname only involves constructing and storing lightweight steering vectors, without gradient computation or optimizer-state storage. Parametric adaptation methods, in contrast, require iterative training and the associated computational and memory overhead for every personalized model update. Together with the high flexibility and robustness validated above, \modelname is a real-world feasible method that offers superior performance gains under scarce and shifting environment.
\vspace{-10pt}

\subsubsection{(RQ6) Hyperparameters Analysis}
\label{sec:hyper}

\begin{figure}[!h]
    \centering
    \includegraphics[width=0.8\linewidth]{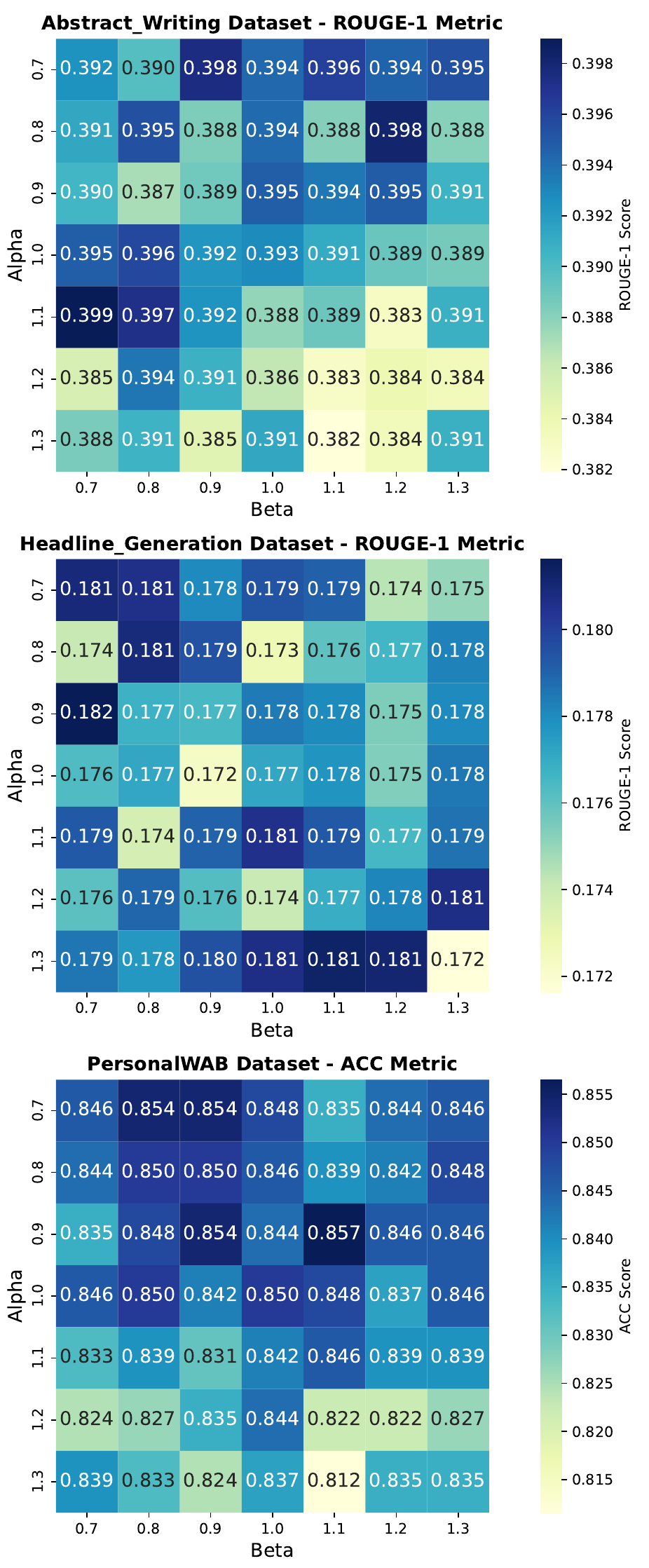}
    \caption{Hyperparameter sensitivity analysis.}
    \label{fig:hyper1}
\end{figure}

We perform grid search on $(\alpha,\beta)$ in a range of $(0.6,1.4)\times(0.6,1.4)$.
The results are visualized in Figure~\ref{fig:hyper1}. Concisely, we compute the variances of the values and find that resulting performances show low variance: 0.000019 on abstract writing, 0.000006 on headline generation, and 0.000092 on PersonalWAB.
 
\begin{figure}[!h]
    \centering
    \includegraphics[width=0.85\linewidth]{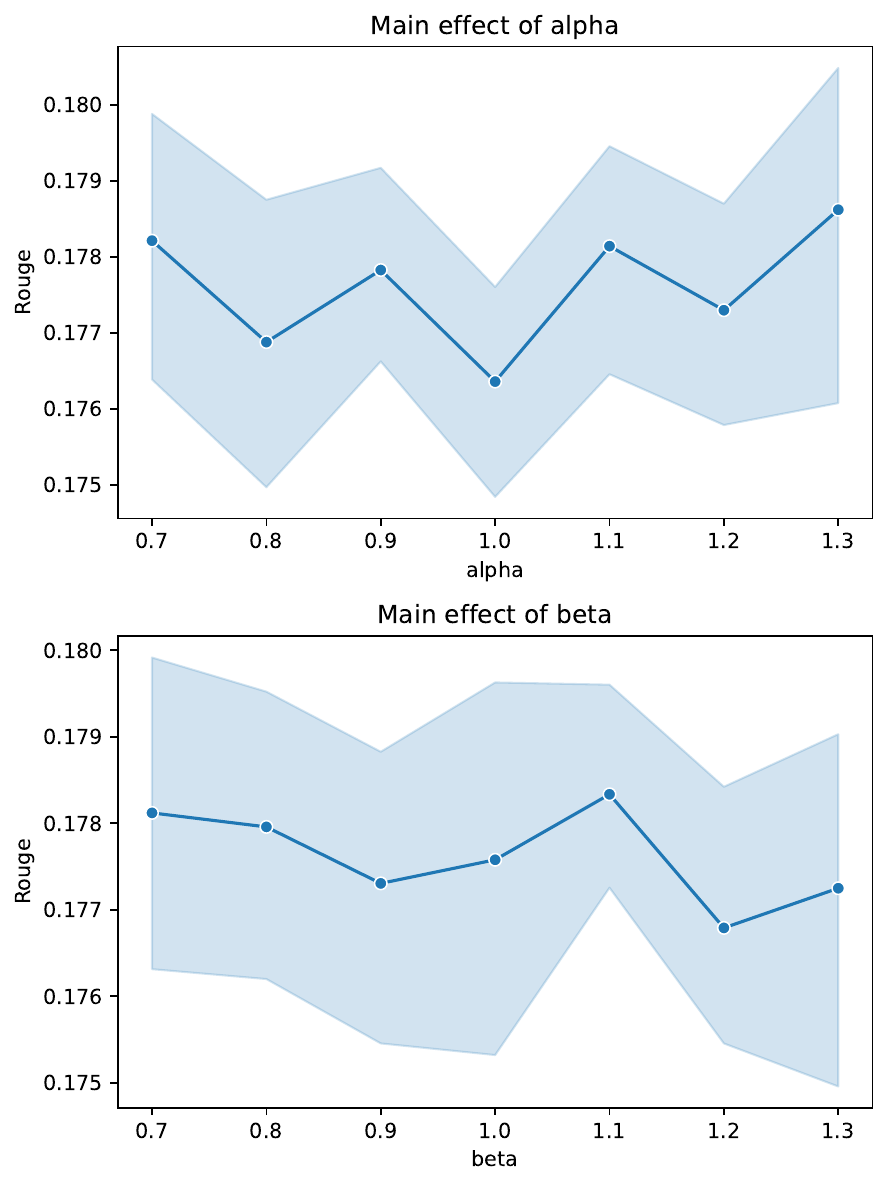}
    \caption{Main effects analysis on Headline Generation.}
    \label{fig:head_main}
\end{figure}
\begin{figure}[!h]
    \centering
    \includegraphics[width=0.85\linewidth]{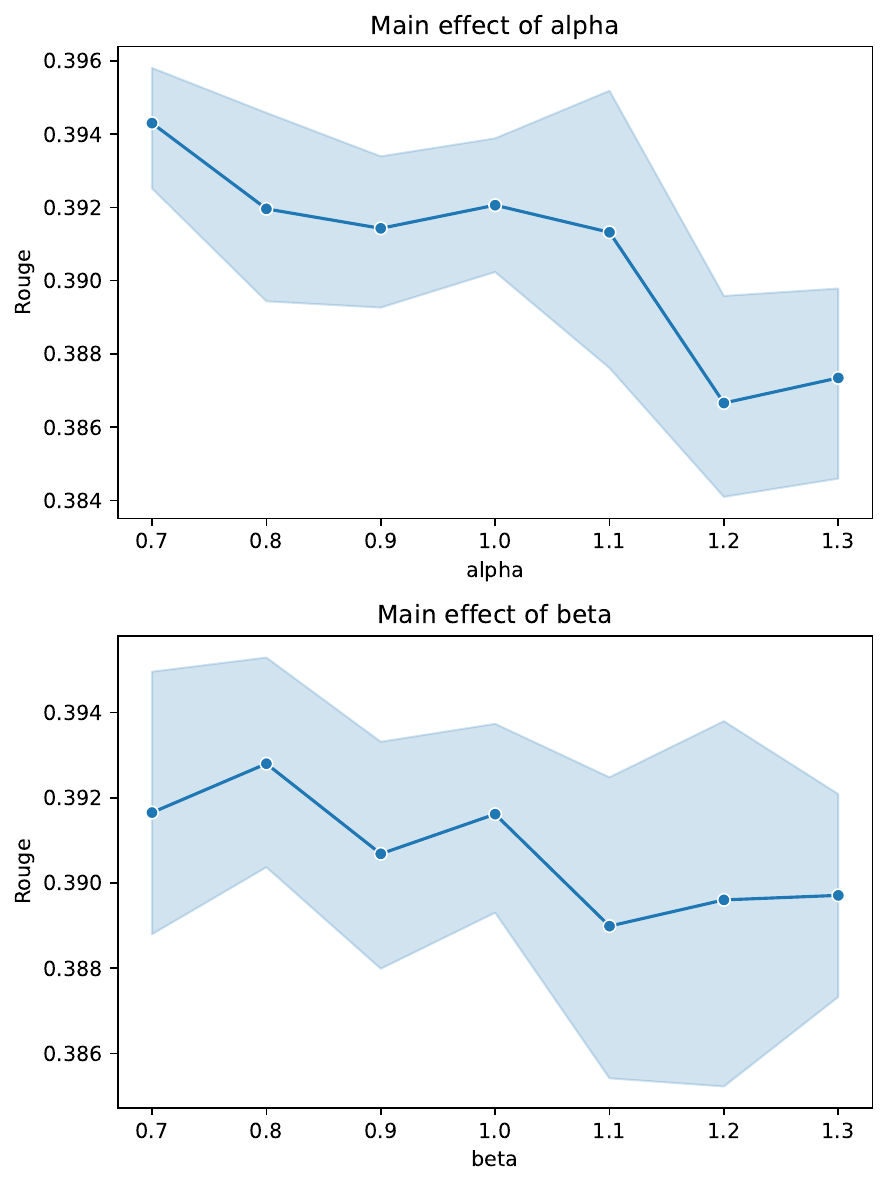}
    \caption{Main effects analysis on Abstract Writing.}
    \label{fig:abs_main}
\end{figure}
\begin{figure}[!h]
    \centering
    \vspace{-10pt}
    \includegraphics[width=0.85\linewidth]{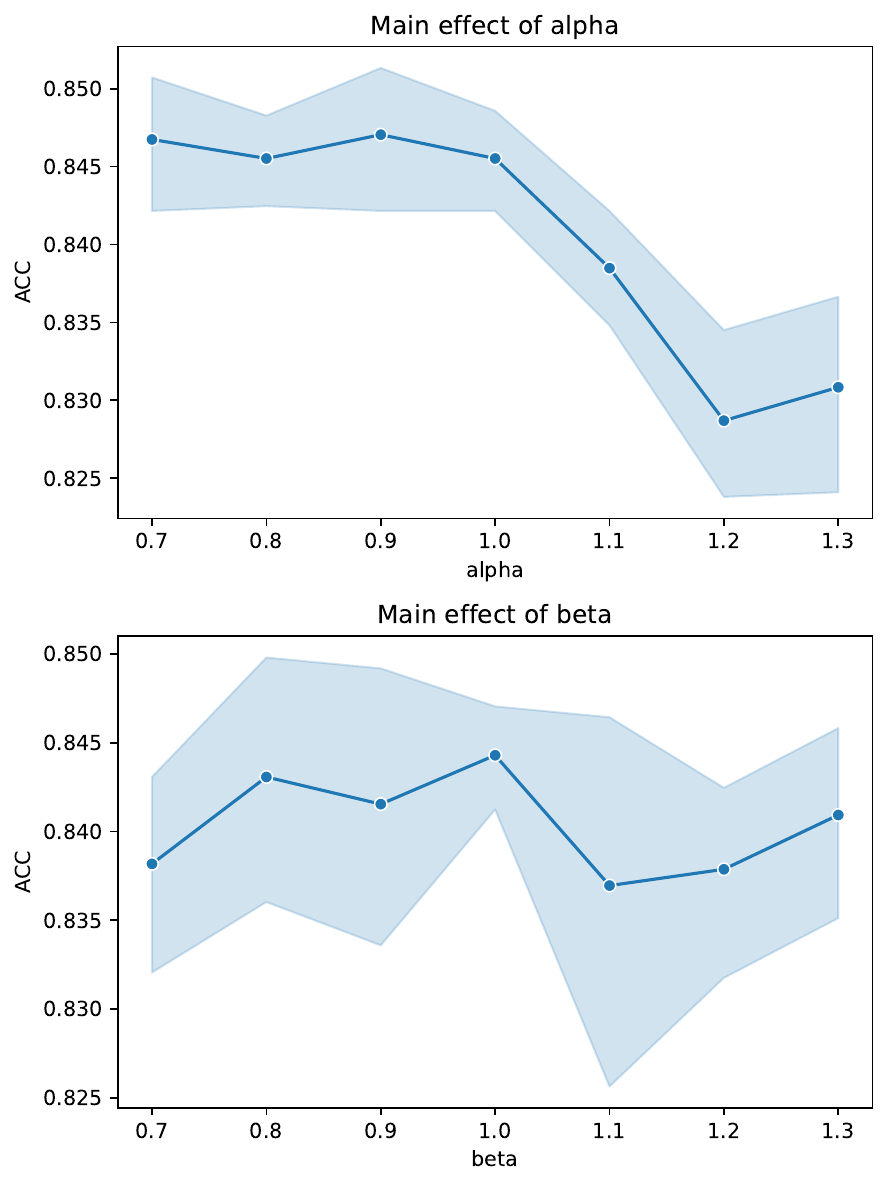}
    \caption{Main effects analysis on PersonalWAB.}
    \label{fig:pwab_main}
\end{figure}
We further conduct one-way/two-way analysis and visualize the main effects analysis in Figure~\ref{fig:head_main}, Figure~\ref{fig:abs_main}, and Figure~\ref{fig:pwab_main}.From the  anova analysis, we can find that $\alpha$ is the only systematically impactful parameter, with $\alpha \le 1.1$ as a general guideline.


\section{Conclusion}
In this paper, we present \modelname, a novel inference-time personalization framework that addresses the critical challenges of data efficiency and adaptability in LLM personalization. Unlike traditional parametric methods that require extensive fine-tuning, \modelname operates through activation steering at inference time, making it particularly suitable for scenarios with fast-changing user preferences and limited training data.
The core innovations of \modelname include: (1) \textit{matched-contrast} that produces user-relevant signals, (2) \textit{fine-grained hooking} that separately extracts steering signals from attention and MLP layers, capturing nuanced user preferences that whole-layer approaches miss; and (3) \textit{query-conditioned aggregation} that dynamically retrieves relevant historical steering vectors based on semantic similarity to the current query, enabling instance-level adaptation without retraining.
Extensive evaluation across diverse tasks demonstrates \modelname's superior performance over both prompt-based and parameter-efficient fine-tuning baselines. Notably, \modelname maintains strong performance even under extreme data scarcity (as few as 5-6 examples per user) and rapidly shifting user distributions, scenarios where traditional methods exhibit significant performance degradation.

\vspace{-6pt}

\begin{acknowledgement}
We thank the anonymous reviewers for their constructive comments and suggestions, which helped us refine the methodological presentation, strengthen the experimental evaluation, and clarify the contributions and limitations of this work.
\end{acknowledgement}


\bibliographystyle{plainnat}
\bibliography{ref}

@inproceedings{rimsky2024steering,
  title     = {Steering {Llama 2} via Contrastive Activation Addition},
  author    = {Rimsky, Nina and Gabrieli, Nick and Schulz, Julian
               and Tong, Meg and Hubinger, Evan and Turner, Alexander},
  booktitle = {Proceedings of the 62nd Annual Meeting of the
               Association for Computational Linguistics
               (Volume 1: Long Papers)},
  year      = {2024},
  pages     = {15504--15522},
  publisher = {Association for Computational Linguistics},
  doi       = {10.18653/v1/2024.acl-long.828}
}

@inproceedings{tan2024democratizing,
  title={Democratizing large language models via personalized parameter-efficient fine-tuning},
  author={Tan, Zhaoxuan and Zeng, Qingkai and Tian, Yijun and Liu, Zheyuan and Yin, Bing and Jiang, Meng},
  booktitle={Proceedings of the 2024 Conference on Empirical Methods in Natural Language Processing},
  pages={6476--6491},
  year={2024}
}

@article{zou2023representation,
  title={Representation engineering: A top-down approach to ai transparency},
  author={Zou, Andy and Phan, Long and Chen, Sarah and Campbell, James and Guo, Phillip and Ren, Richard and Pan, Alexander and Yin, Xuwang and Mazeika, Mantas and Dombrowski, Ann-Kathrin and others},
  journal={arXiv preprint arXiv:2310.01405},
  year={2023}
}

@article{chen2025persona,
  title={Persona vectors: Monitoring and controlling character traits in language models},
  author={Chen, Runjin and Arditi, Andy and Sleight, Henry and Evans, Owain and Lindsey, Jack},
  journal={arXiv preprint arXiv:2507.21509},
  year={2025}
}

@inproceedings{ryan2025synthesizeme,
  title={SynthesizeMe! inducing persona-guided prompts for personalized reward models in LLMs},
  author={Ryan, Michael J and Shaikh, Omar and Bhagirath, Aditri and Frees, Daniel and Held, William Barr and Yang, Diyi},
  booktitle={Proceedings of the 63rd Annual Meeting of the Association for Computational Linguistics (Volume 1: Long Papers)},
  pages={8045--8078},
  year={2025}
}

@inproceedings{banerjee2005meteor,
  title={METEOR: An automatic metric for MT evaluation with improved correlation with human judgments},
  author={Banerjee, Satanjeev and Lavie, Alon},
  booktitle={Proceedings of the acl workshop on intrinsic and extrinsic evaluation measures for machine translation and/or summarization},
  pages={65--72},
  year={2005}
}

@article{zhang2025personalized,
  title={Personalized Text Generation with Contrastive Activation Steering},
  author={Zhang, Jinghao and Liu, Yuting and Wang, Wenjie and Liu, Qiang and Wu, Shu and Wang, Liang and Chua, Tat-Seng},
  journal={arXiv preprint arXiv:2503.05213},
  year={2025}
}

@article{vuong2021does,
  title={Does more context help? Effects of context window and application source on retrieval performance},
  author={Vuong, Tung and Andolina, Salvatore and Jacucci, Giulio and Ruotsalo, Tuukka},
  journal={ACM Transactions on Information Systems (TOIS)},
  volume={40},
  number={2},
  pages={1--40},
  year={2021},
  publisher={ACM New York, NY}
}

@inproceedings{guha2015user,
  title={User modeling for a personal assistant},
  author={Guha, Ramanathan and Gupta, Vineet and Raghunathan, Vivek and Srikant, Ramakrishnan},
  booktitle={Proceedings of the Eighth ACM International Conference on Web Search and Data Mining},
  pages={275--284},
  year={2015}
}

@article{hu2022lora,
  title={Lora: Low-rank adaptation of large language models.},
  author={Hu, Edward J and Shen, Yelong and Wallis, Phillip and Allen-Zhu, Zeyuan and Li, Yuanzhi and Wang, Shean and Wang, Lu and Chen, Weizhu and others},
  journal={ICLR},
  volume={1},
  number={2},
  pages={3},
  year={2022}
}

@article{qian2024memorag,
  title={Memorag: Moving towards next-gen rag via memory-inspired knowledge discovery},
  author={Qian, Hongjin and Zhang, Peitian and Liu, Zheng and Mao, Kelong and Dou, Zhicheng},
  journal={arXiv preprint arXiv:2409.05591},
  volume={1},
  year={2024}
}

@article{dong2022survey,
  title={A survey on in-context learning},
  author={Dong, Qingxiu and Li, Lei and Dai, Damai and Zheng, Ce and Ma, Jingyuan and Li, Rui and Xia, Heming and Xu, Jingjing and Wu, Zhiyong and Liu, Tianyu and others},
  journal={arXiv preprint arXiv:2301.00234},
  year={2022}
}

@article{gui2024bonbon,
  title={Bonbon alignment for large language models and the sweetness of best-of-n sampling},
  author={Gui, Lin and G{\^a}rbacea, Cristina and Veitch, Victor},
  journal={Advances in Neural Information Processing Systems},
  volume={37},
  pages={2851--2885},
  year={2024}
}

@article{liu2023statistical,
  title={Statistical rejection sampling improves preference optimization},
  author={Liu, Tianqi and Zhao, Yao and Joshi, Rishabh and Khalman, Misha and Saleh, Mohammad and Liu, Peter J and Liu, Jialu},
  journal={arXiv preprint arXiv:2309.06657},
  year={2023}
}

@article{zhang2025cold,
  title={Cold-start recommendation towards the era of large language models (llms): A comprehensive survey and roadmap},
  author={Zhang, Weizhi and Bei, Yuanchen and Yang, Liangwei and Zou, Henry Peng and Zhou, Peilin and Liu, Aiwei and Li, Yinghui and Chen, Hao and Wang, Jianling and Wang, Yu and others},
  journal={arXiv preprint arXiv:2501.01945},
  year={2025}
}

@article{amazon2025context,
  title={Context Length Alone Hurts LLM Performance Despite Perfect Retrieval},
  author={Amazon Science and Research Teams},
  journal={Amazon Science Publications},
  year={2025},
  url={https://www.amazon.science/publications/context-length-alone-hurts-llm-performance-despite-perfect-retrieval}
}

@inproceedings{liu2024retrieval,
  title={Relia: Retrieval-enhanced Large Language Model for Lifelong Sequential Behavior Comprehension in Recommendation},
  author={Liu, Jianghao and Shan, Rong and Zhu, Chenxu and Du, Kounanhua and Chen, Bo and Quan, Shigang and Tang, Ruiming and Yu, Yong and Zhang, Weinan},
  booktitle={Proceedings of the ACM Web Conference 2024},
  pages={3497--3508},
  year={2024}
}

@article{jiang2024importance,
  title={Importance Weighting Can Help Large Language Models Self-Improve},
  author={Jiang, Yifan and others},
  journal={arXiv preprint arXiv:2408.09849},
  year={2024}
}

@inproceedings{lin2004rouge,
  title={Rouge: A package for automatic evaluation of summaries},
  author={Lin, Chin-Yew},
  booktitle={Text summarization branches out},
  pages={74--81},
  year={2004}
}

@article{saleem2025ai,
  title={AI in education: Personalized learning systems and their impact on student performance and engagement},
  author={Saleem, Sadaf and Aziz, Muhammad Umar and Iqbal, Muhammad Jawed and Abbas, Shahid},
  journal={The Critical Review of Social Sciences Studies},
  volume={3},
  number={1},
  pages={2445--2459},
  year={2025}
}

@article{iqbal2025personalized,
  title={Personalized and adaptive e-learning systems for semantic Web: a systematic review and roadmap},
  author={Iqbal, Muddesar and Sarwar, Sohail and Safyan, Muhammad and Nasralla, Moustafa},
  journal={International Journal of Web Information Systems},
  volume={21},
  number={4},
  pages={327--352},
  year={2025},
  publisher={Emerald Publishing Limited}
}

@inproceedings{liu2025proactive,
  title={Proactive conversational agents with inner thoughts},
  author={Liu, Xingyu Bruce and Fang, Shitao and Shi, Weiyan and Wu, Chien-Sheng and Igarashi, Takeo and Chen, Xiang'Anthony'},
  booktitle={Proceedings of the 2025 CHI Conference on Human Factors in Computing Systems},
  pages={1--19},
  year={2025}
}

@article{yusuf2025pedagogical,
  title={Pedagogical AI conversational agents in higher education: a conceptual framework and survey of the state of the art},
  author={Yusuf, Habeeb and Money, Arthur and Daylamani-Zad, Damon},
  journal={Educational technology research and development},
  volume={73},
  number={2},
  pages={815--874},
  year={2025},
  publisher={Springer}
}

@article{wang2025roles,
  title={Roles of artificial intelligence experience, information redundancy, and familiarity in shaping active learning: Insights from intelligent personal assistants},
  author={Wang, Shaofeng and Sun, Zhuo},
  journal={Education and Information Technologies},
  volume={30},
  number={2},
  pages={2525--2546},
  year={2025},
  publisher={Springer}
}

@article{turner2023steering,
  title   = {Steering Language Models with Activation Engineering},
  author  = {Turner, Alexander M and Thiergart, Laura and Leech, Gabriel and Udell, David and Vazquez, Juan J and Mini, Ulisse and MacDiarmid, Monte},
  journal = {arXiv preprint arXiv:2308.10248},
  year    = {2023}
}

@inproceedings{frandsen2022extracting,
  title     = {Extracting Latent State Representations with Linear Dynamics from Language Models},
  author    = {Frandsen, Abraham and Ge, Rong and Sharan, Vatsal},
  booktitle = {Proc. of ICML},
  year      = {2022}
}

@inproceedings{lin2024rella,
  title={Rella: Retrieval-enhanced large language models for lifelong sequential behavior comprehension in recommendation},
  author={Lin, Jianghao and Shan, Rong and Zhu, Chenxu and Du, Kounianhua and Chen, Bo and Quan, Shigang and Tang, Ruiming and Yu, Yong and Zhang, Weinan},
  booktitle={Proceedings of the ACM Web Conference 2024},
  pages={3497--3508},
  year={2024}
}

@article{liu2025survey,
  title={A survey of personalized large language models: Progress and future directions},
  author={Liu, Jiahong and Qiu, Zexuan and Li, Zhongyang and Dai, Quanyu and Zhu, Jieming and Hu, Minda and Yang, Menglin and King, Irwin},
  journal={arXiv preprint arXiv:2502.11528},
  year={2025}
}

@article{yang2025qwen3,
  title={Qwen3 technical report},
  author={Yang, An and Li, Anfeng and Yang, Baosong and Zhang, Beichen and Hui, Binyuan and Zheng, Bo and Yu, Bowen and Gao, Chang and Huang, Chengen and Lv, Chenxu and others},
  journal={arXiv preprint arXiv:2505.09388},
  year={2025}
}

@article{dubey2024llama,
  title={The llama 3 herd of models},
  author={Dubey, Abhimanyu and Jauhri, Abhinav and Pandey, Abhinav and Kadian, Abhishek and Al-Dahle, Ahmad and Letman, Aiesha and Mathur, Akhil and Schelten, Alan and Yang, Amy and Fan, Angela and others},
  journal={arXiv e-prints},
  pages={arXiv--2407},
  year={2024}
}

@inproceedings{cai2025large,
  title={Large language models empowered personalized web agents},
  author={Cai, Hongru and Li, Yongqi and Wang, Wenjie and Zhu, Fengbin and Shen, Xiaoyu and Li, Wenjie and Chua, Tat-Seng},
  booktitle={Proceedings of the ACM on Web Conference 2025},
  pages={198--215},
  year={2025}
}

@article{kumar2024longlamp,
  title={Longlamp: A benchmark for personalized long-form text generation},
  author={Kumar, Ishita and Viswanathan, Snigdha and Yerra, Sushrita and Salemi, Alireza and Rossi, Ryan A and Dernoncourt, Franck and Deilamsalehy, Hanieh and Chen, Xiang and Zhang, Ruiyi and Agarwal, Shubham and others},
  journal={arXiv preprint arXiv:2407.11016},
  year={2024}
}

@article{misra2022news,
  title={News category dataset},
  author={Misra, Rishabh},
  journal={arXiv preprint arXiv:2209.11429},
  year={2022}
}

@article{salemi2023lamp,
  title={Lamp: When large language models meet personalization},
  author={Salemi, Alireza and Mysore, Sheshera and Bendersky, Michael and Zamani, Hamed},
  journal={arXiv preprint arXiv:2304.11406},
  year={2023}
}

@article{wang2025never,
  title={Never Start from Scratch: Expediting On-Device LLM Personalization via Explainable Model Selection},
  author={Wang, Haoming and Yang, Boyuan and Yin, Xiangyu and Gao, Wei},
  journal={arXiv preprint arXiv:2504.13938},
  year={2025}
}

@article{madaan2022memory,
  title={Memory-assisted prompt editing to improve GPT-3 after deployment},
  author={Madaan, Aman and Tandon, Niket and Clark, Peter and Yang, Yiming},
  journal={arXiv preprint arXiv:2201.06009},
  year={2022}
}

@article{zhuang2024hydra,
  title={Hydra: Model factorization framework for black-box llm personalization},
  author={Zhuang, Yuchen and Sun, Haotian and Yu, Yue and Qiang, Rushi and Wang, Qifan and Zhang, Chao and Dai, Bo},
  journal={Advances in Neural Information Processing Systems},
  volume={37},
  pages={100783--100815},
  year={2024}
}

@article{tan2024personalized,
  title={Personalized pieces: Efficient personalized large language models through collaborative efforts},
  author={Tan, Zhaoxuan and Liu, Zheyuan and Jiang, Meng},
  journal={arXiv preprint arXiv:2406.10471},
  year={2024}
}

@inproceedings{CFRAG,
author = {Shi, Teng and Xu, Jun and Zhang, Xiao and Zang, Xiaoxue and Zheng, Kai and Song, Yang and Li, Han},
title = {Retrieval Augmented Generation with Collaborative Filtering for Personalized Text Generation},
year = {2025},
isbn = {9798400715921},
publisher = {Association for Computing Machinery},
address = {New York, NY, USA},
url = {https://doi.org/10.1145/3726302.3730075},
doi = {10.1145/3726302.3730075},
booktitle = {Proceedings of the 48th International ACM SIGIR Conference on Research and Development in Information Retrieval},
pages = {1294–1304},
numpages = {11},
location = {Padua, Italy},
series = {SIGIR '25}
}

@misc{PLoRA,
      title={Personalized LoRA for Human-Centered Text Understanding}, 
      author={You Zhang and Jin Wang and Liang-Chih Yu and Dan Xu and Xuejie Zhang},
      year={2024},
      eprint={2403.06208},
      archivePrefix={arXiv},
      primaryClass={cs.CL},
      url={https://arxiv.org/abs/2403.06208}, 
}

@inproceedings{
chen2025pal,
title={{PAL}: Sample-Efficient Personalized Reward Modeling for Pluralistic Alignment},
author={Daiwei Chen and Yi Chen and Aniket Rege and Zhi Wang and Ramya Korlakai Vinayak},
booktitle={The Thirteenth International Conference on Learning Representations},
year={2025},
url={https://openreview.net/forum?id=1kFDrYCuSu}
}


\end{document}